\documentclass{article}
\usepackage{spconf,amsmath,graphicx,hyperref}
\usepackage{multirow}
\usepackage[bottom,flushmargin,hang,multiple]{footmisc}

\title{ARMA Block: A CNN-Based Autoregressive and Moving Average Module for Long-Term Time Series Forecasting}
\name{Myung Jin Kim$^{1,2}$, YeongHyeon Park$^{3}$, Il Dong Yun$^{2}$\thanks{This research was supported by Grepp.\\ Corresponding author: Il Dong Yun.}}
\address{$^{1}$AI Lab, Grepp\\ $^{2}$Department of Computer Science and Engineering, Hankuk University of Foreign Studies \\ $^{3}$Department of Radiation Physics, The University of Texas MD Anderson Cancer Center \\
\texttt{jin@grepp.co, ypark6@mdanderson.org, yun@hufs.ac.kr}}
\begin{document}
%\ninept
%
\maketitle
\begin{abstract}
This paper proposes a simple yet effective convolutional module for long-term time series forecasting. The proposed block, inspired by the Auto-Regressive Integrated Moving Average (ARIMA) model, consists of two convolutional components: one for capturing the trend (autoregression) and the other for refining local variations (moving average). Unlike conventional ARIMA, which requires iterative multi-step forecasting, the block directly performs multi-step forecasting, making it easily extendable to multivariate settings. Experiments on nine widely used benchmark datasets demonstrate that our method ARMA achieves competitive accuracy, particularly on datasets exhibiting strong trend variations, while maintaining architectural simplicity. Furthermore, analysis shows that the block inherently encodes absolute positional information, suggesting its potential as a lightweight replacement for positional embeddings in sequential models.
\end{abstract}
\section{Introduction}
\label{sec:intro}
Recently, the Time Series Forecasting (TSF)  has become an increasingly important area. Time series forecasting has moved beyond simply predicting weather or traffic, and is being used to learn patterns from any data that has an order to produce results. For example, the Transformer \cite{vaswani2017attention} has been successful in the field of natural language processing by detecting patterns in sequences and making predictions about the output to create linguistic prediction models, and the Vision Transformer \cite{dosovitskiy2020image} is being used as a fundamental model to better understand the characteristics of images by using individual patches of an image as time series data. Recently, Mamba \cite{gu2023mamba, gu2024mamba} attempted to make existing transformers more effective using structured state space models (SSM). This suggests that extending the problem to better predict time series can enhance the efficiency and performance of current state-of-the-art models.

Based on the successful application of transformers in various scenarios, numerous models employing transformers are now being proposed within the TSF domain\cite{zhou2021informer, wu2021autoformer, zhou2022fedformer}. Recently, the introduction of transformers into TFS has become central, including the application of models utilising Mixture of Experts (MoE)\cite{shi2024time, liu2024mixture}, which represents one direction in the advancement of Large Language Models (LLMs) by achieving both efficiency and performance.

However, amidst this deluge of transformer models, the emergence of the remarkably simple DLinear \cite{zeng2023transformers} model caused quite a stir. This model, combining a straightforward architecture with linear layers, achieved exceptionally high TSF accuracy, demonstrating that simpler models can outperform complex transformer-based ones.

Meanwhile, it has been discovered that Convolutional Neural Networks (CNNs) learn absolute positional information during training \cite{islam2020much}. Models such as Transformers perform parallel processing and therefore cannot directly learn positional information from features; consequently, positional embeddings are added separately. However, The CNNs learn information about absolute positions from padding when learning each feature. This can be considered an inherent characteristic of CNNs.

Therefore, this paper proposes a model that performs time series forecasting alongside absolute position information. This model is highly simple and can be easily integrated into various models through modularisation. Based on the description of the proposed model architecture, this paper evaluates the data using the most popular LTFS dataset, which is primarily employed for time series forecasting assessment. Also, it demonstrates that the trained modules incorporate absolute positional information.

\section{Method}
\label{sec:method}

\begin{figure}[htb]
\centering
\includegraphics[width=4.5cm]{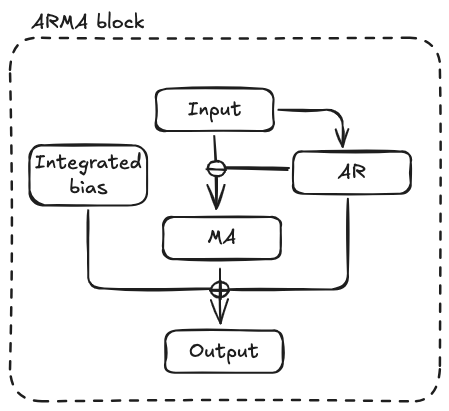}
\caption{Block diagram of ARMA block.}
\label{fig:arima_block}
\end{figure}

This section explains the structure of the Auto Regressive Moving Average (ARMA) block. The block diagram is shown in Fig \ref{fig:arima_block}. The block is simple, consisting of two CNN components and a bias. Each part predicts the trend and details separately. Although it appears structurally similar to DLinear, the underlying mechanism is distinct. It was created inspired by the ARIMA model. While conventional ARIMA models perform univariate forecasting and incorporate an integration step to correct errors through iterative multi-step forecasting, this block has been configured to perform only direct multiple-step forecasting. This modification facilitates its extension to multivariate applications. 
In ARIMA, the method employed involves repeatedly making forecasts to reduce errors in the short term. However, the ARMA block enhances efficiency by directly forecasting multiple steps, using a high-dimensional filter to predict the characteristics of each step.

The ARMA block's operations can be expressed as below.

\begin{equation}
    y_{ar} = AR(x_{in})
\label{eq:AR}
\end{equation}
\begin{equation}
    y_{ma} = MA(x_{in} - y_{ar})
\label{eq:MA}
\end{equation}
\begin{equation}
    y_{out} = y_{ar} + y_{ma}
\label{eq:last}
\end{equation}

We shall now proceed to describe each CNN layer constituting the ARMA block.

\subsection{Auto-Regression (AR) prediction}
\label{ssec:AR_sec}

The AR component predicts the trend of predicted data. As seen in \eqref{eq:AR}, the AR component receives the initial input, while the MA component takes the value generated by the AR and subtracts it from the input value before processing. Due to the optimiser's nature, learning proceeds in the direction of the larger gradient, meaning the AR component learns the overall trend.

\subsection{Moving Average (MA) prediction}
\label{ssec:MA_sec}

The MA component refines the details. The layer receives all detail values from the input except for the trend value, and is responsible for performing detailed trend correction on the final output signal. As the input values are those with the trend removed, they are likely to be in the higher frequency range. Since it predicts these, this layer is responsible for the detailed aspects in the final prediction.

\begin{figure}[htb]
\includegraphics[width=8.5cm]{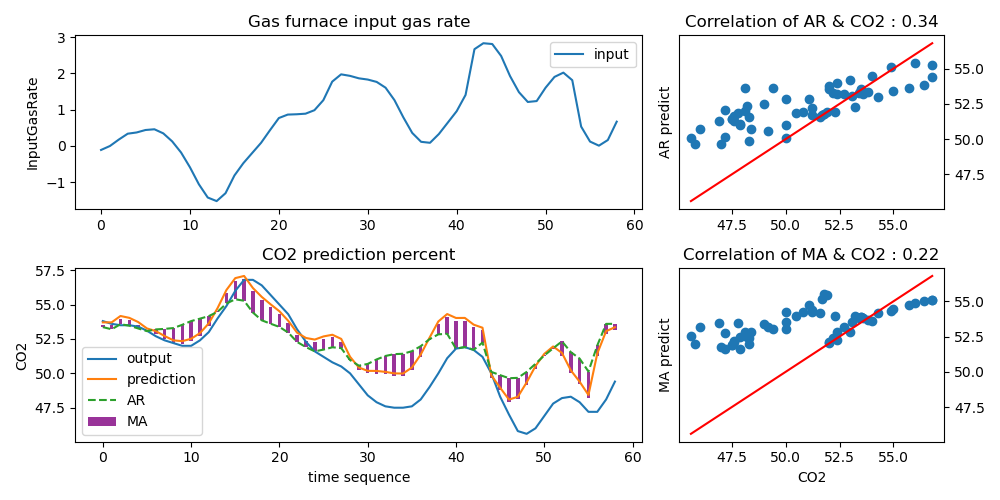}
\caption{The ARMA block prediction example. The AR part predicts the CO2 trend, while the MA part refines the residual details.}
\label{fig:arima_example}
\end{figure}

Fig. \ref{fig:arima_example}. Qualitative illustration of the ARMA block prediction. This gas furnace time series data, from Time Series Analysis \cite{box2015time}, uses the gas rate as input to predict CO2 percentage. The orange line (final output) follows the CO2 trend well. The AR layer (green dotted line) partially reflects the inverse relation between input and output, while the MA layer (purple bars) corrects detailed residual errors. Correlation values show the AR component (0.34) exceeds the MA (0.22), suggesting the AR portion learns the trend of CO2.

\section{Experiments}
\label{sec:exp}

\subsection{Experiments Settings}
\label{ssec:environments}

\textbf{Dataset.} We conduct extensive experiments on nine widely-used multivariate real-world datasets, including ETT (Electricity Transformer Temperature) \cite{zhou2021informer} (ETTh1, ETTh2, ETTm1, ETTm2), Traffic, Electricity, Weather, ILI, ExchangeRate \cite{lai2018modeling}. All of them are often used multivariate time series.

\textbf{Training environments.} We adopt a standard training setup. The batch size is 32. The learning rate is set to 0.001, and the AdamW optimizer is employed. Additionally, we use the RevIN \cite{kim2021reversible} module to improve robustness against non-stationarity. The ARMA block allows adjustment of the filter size for both AR and MA components; in this experiment, a 5x5 filter was employed.

\textbf{Evaluation metrics.} We use Mean Square Error (MSE) and Mean Absolute Error (MAE) for evaluation metric. These metrics are generally used \cite{zhou2021informer, wu2021autoformer, zeng2023transformers, shi2024time}.

%\subsection{Experiments data}
%\label{ssec:data}

\begin{table*}[h!]
\begin{center}
\caption{The prediction errors of the nine popular datasets for the LTSF problem.($^{*}$ with SAN \cite{liu2023adaptive}, best only)}
\label{table:result}
\begin{tabular}{ c|c|c|c|c|c|c|c|c|c|c|c|c|c } 
 \hline
 \multicolumn{2}{c|}{Methods} & \multicolumn{2}{c|}{Ours} & \multicolumn{2}{c|}{NLinear} & \multicolumn{2}{c|}{DLinear} & \multicolumn{2}{c|}{FEDformer} & \multicolumn{2}{c|}{Autoformer} & \multicolumn{2}{c}{Informer} \\ 
 \hline
 \multicolumn{2}{c|}{Metric} & MSE & MAE & MSE & MAE & MSE & MAE & MSE & MAE & MSE & MAE & MSE & MAE \\
 \hline
 \multirow{4}{*}{\rotatebox{90}{Electricity}} & 96 & 0.206 & 0.299 & 0.141 & \textbf{0.237} & \textbf{0.137$^{*}$} & \textbf{0.234$^{*}$} & 0.164$^{*}$ & 0.272$^{*}$ & 0.172$^{*}$ & 0.281$^{*}$ & 0.274 & 0.368 \\
 & 192 & 0.209 & 0.301 & 0.154 & 0.248 & \textbf{0.151$^{*}$} & \textbf{0.247$^{*}$} & 0.179$^{*}$ & 0.286$^{*}$ & 0.195$^{*}$ & 0.300$^{*}$ & 0.296 & 0.386 \\
 & 336 & 0.228 & 0.319 & 0.171 & 0.265 & \textbf{0.166$^{*}$} & \textbf{0.264$^{*}$} & 0.191$^{*}$ & 0.299$^{*}$ & 0.211$^{*}$ & 0.316$^{*}$ & 0.300 & 0.394 \\
 & 720 & 0.261 & 0.344 & 0.210 & 0.297 & \textbf{0.201$^{*}$} & \textbf{0.295$^{*}$} & 0.230$^{*}$ & 0.334$^{*}$ & 0.236$^{*}$ & 0.335$^{*}$ & 0.373 & 0.439 \\
 \hline
 \multirow{4}{*}{\rotatebox{90}{Exchange}} & 96 & \textbf{0.059} & \textbf{0.175} & 0.089 & 0.208 & 0.081 & 0.203 & 0.079$^{*}$ & 0.205$^{*}$ & 0.082$^{*}$ & 0.208$^{*}$ & 0.847 & 0.752 \\
 & 192 & \textbf{0.114} & \textbf{0.257} & 0.180 & 0.300 & 0.157 & 0.293 & 0.156$^{*}$ & 0.295$^{*}$ & 0.157$^{*}$ & 0.296$^{*}$ & 1.204 & 0.895 \\
 & 336 & \textbf{0.213} & \textbf{0.340} & 0.331 & 0.415 & 0.294$^{*}$ & 0.407$^{*}$ & 0.260$^{*}$ & 0.384$^{*}$ & 0.262$^{*}$ & 0.385 & 1.672 & 10.36 \\
 & 720 & \textbf{0.581} & \textbf{0.584} & 1.033 & 0.780 & 0.643 & 0.601 & 0.697$^{*}$ & 0.633$^{*}$ & 0.689$^{*}$ & 0.629 & 2.478 & 1.310 \\
 \hline
 \multirow{4}{*}{\rotatebox{90}{Traffic}} & 96 & 0.565 & 0.433 & \textbf{0.410} & \textbf{0.279} & \textbf{0.410} & 0.282 & 0.536$^{*}$ & 0.330$^{*}$ & 0.569$^{*}$ & 0.350$^{*}$ & 0.719 & 0.391 \\
 & 192 & 0.513 & 0.399 & \textbf{0.423} & \textbf{0.284} & \textbf{0.423} & 0.287 & 0.536$^{*}$ & 0.330$^{*}$ & 0.569$^{*}$ & 0.350 & 0.696 & 0.379 \\
 & 336 & 0.548 & 0.415 & \textbf{0.435} & \textbf{0.290} & 0.436 & 0.296 & 0.580$^{*}$ & 0.354$^{*}$ & 0.591$^{*}$ & 0.363 & 0.777 & 0.420 \\
 & 720 & 0.526 & 0.398 & \textbf{0.464} & \textbf{0.307} & 0.466 & 0.315 & 0.607$^{*}$ & 0.367$^{*}$ & 0.623$^{*}$ & 0.380 & 0.864 & 0.472 \\
 \hline
 \multirow{4}{*}{\rotatebox{90}{Weather}} & 96 & 0.230 & 0.269 & 0.182 & 0.232 & \textbf{0.152$^{*}$} & \textbf{0.210$^{*}$} & 0.179$^{*}$ & 0.239$^{*}$ & 0.194$^{*}$ & 0.256$^{*}$ & 0.300 & 0.384 \\
 & 192 & 0.293 & 0.317 & 0.225 & 0.269 & \textbf{0.196$^{*}$} & \textbf{0.254$^{*}$} & 0.234$^{*}$ & 0.296$^{*}$ & 0.258$^{*}$ & 0.316$^{*}$ & 0.598 & 0.544 \\
 & 336 & 0.353 & 0.358 & 0.271 & 0.301 & \textbf{0.246$^{*}$} & \textbf{0.294$^{*}$} & 0.304$^{*}$ & 0.348$^{*}$ & 0.329$^{*}$ & 0.367$^{*}$ & 0.578 & 0.523 \\
 & 720 & 0.446 & 0.415 & 0.338 & 0.348 & \textbf{0.315$^{*}$} & \textbf{0.346$^{*}$} & 0.400$^{*}$ & 0.404$^{*}$ & 0.419 & 0.428 & 1.059 & 0.741 \\
 \hline
 \multirow{4}{*}{\rotatebox{90}{ILI}} & 24 & \textbf{1.599} & \textbf{0.833} & 1.683 & 0.858 & 2.122$^{*}$ & 1.001$^{*}$ & 2.614$^{*}$ & 1.119$^{*}$ & 2.777$^{*}$ & 1.157$^{*}$ & 5.764 & 1.677 \\
 & 36 & \textbf{1.353} & \textbf{0.776} & 1.703 & 0.869 & 1.963 & 0.963 & 2.537$^{*}$ & 1.079$^{*}$ & 2.649$^{*}$ & 1.104$^{*}$ & 4.755 & 1.467 \\
 & 48 & \textbf{1.422} & \textbf{0.792} & 1.719 & 0.884 & 2.041$^{*}$ & 0.971$^{*}$ & 2.416$^{*}$ & 1.032$^{*}$ & 2.420$^{*}$ & 1.029$^{*}$ & 4.763 & 1.469 \\
 & 60 & \textbf{1.435} & \textbf{0.782} & 1.819 & 0.917 & 2.089$^{*}$ & 0.973$^{*}$ & 2.299$^{*}$ & 1.003$^{*}$ & 2.401$^{*}$ & 1.021$^{*}$ & 5.264 & 1.564 \\
 \hline
 \multirow{4}{*}{\rotatebox{90}{ETTh1}} & 96 & 0.436 & 0.458 & \textbf{0.374} & \textbf{0.394} & 0.375 & 0.399 & 0.376 & 0.419 & 0.449 & 0.459 & 0.865 & 0.713 \\
 & 192 & 0.494 & 0.495 & 0.408 & \textbf{0.415} & \textbf{0.405} & 0.416 & 0.420 & 0.448 & 0.500 & 0.482 & 1.008 & 0.792 \\
 & 336 & 0.551 & 0.531 & \textbf{0.429} & \textbf{0.427} & 0.439 & 0.443 & 0.459 & 0.465 & 0.521 & 0.496 & 1.107 & 0.809 \\
 & 720 & 0.718 & 0.626 & \textbf{0.440} & \textbf{0.453} & 0.472 & 0.490 & 0.506 & 0.507 & 0.514 & 0.512 & 1.181 & 0.865 \\
 \hline
 \multirow{4}{*}{\rotatebox{90}{ETTh2}} & 96 & \textbf{0.165} & \textbf{0.277} & 0.277 & 0.338 & 0.277$^{*}$ & 0.338$^{*}$ & 0.300$^{*}$ & 0.355$^{*}$ & 0.316$^{*}$ & 0.366$^{*}$ & 0.865 & 0.713 \\
 & 192 & \textbf{0.191} & \textbf{0.301} & 0.344 & 0.381 & 0.340$^{*}$ & 0.378$^{*}$ & 0.392$^{*}$ & 0.413$^{*}$ & 0.413$^{*}$ & 0.426$^{*}$ & 5.602 & 1.931 \\
 & 336 & \textbf{0.212} & \textbf{0.319} & 0.357 & 0.400 & 0.356$^{*}$ & 0.398$^{*}$ & 0.459$^{*}$ & 0.462$^{*}$ & 0.446$^{*}$ & 0.457$^{*}$ & 4.721 & 1.835 \\
 & 720 & \textbf{0.261} & \textbf{0.356} & 0.394 & 0.436 & 0.396$^{*}$ & 0.435$^{*}$ & 0.462$^{*}$ & 0.472$^{*}$ & 0.471$^{*}$ & 0.474$^{*}$ & 3.647 & 1.625 \\
 \hline
 \multirow{4}{*}{\rotatebox{90}{ETTm1}} & 96 & 0.339 & 0.381 & 0.306 & 0.348 & \textbf{0.299} & \textbf{0.343} & 0.379 & 0.419 & 0.505 & 0.475 & 0.672 & 0.571\\
 & 192 & 0.401 & 0.410 & 0.349 & 0.375 & \textbf{0.335} & \textbf{0.365} & 0.426 & 0.441 & 0.553 & 0.496 & 0.795 & 0.669 \\
 & 336 & 0.468 & 0.448 & 0.375 & 0.388 & \textbf{0.369} & \textbf{0.386} & 0.445 & 0.459 & 0.621 & 0.537 & 1.212 & 0.871 \\
 & 720 & 0.543 & 0.498 & 0.433 & 0.422 & \textbf{0.425} & \textbf{0.421} & 0.543 & 0.490 & 0.671 & 0.561 & 1.166 & 0.823 \\
 \hline
 \multirow{4}{*}{\rotatebox{90}{ETTm2}} & 96 & \textbf{0.113} & \textbf{0.224} & 0.167 & 0.255 & 0.167 & 0.260 & 0.203 & 0.287 & 0.255 & 0.339 & 0.365 & 0.453 \\
 & 192 & \textbf{0.139} & \textbf{0.250} & 0.221 & 0.293 & 0.224 & 0.303 & 0.269 & 0.328 & 0.281 & 0.340 & 0.533 & 0.563 \\
 & 336 & \textbf{0.171} & \textbf{0.277} & 0.274 & 0.327 & 0.281 & 0.342 & 0.325 & 0.366 & 0.339 & 0.372 & 1.363 & 0.887 \\
 & 720 & \textbf{0.222} & \textbf{0.317} & 0.368 & 0.384 & 0.397 & 0.421 & 0.421 & 0.415 & 0.433 & 0.432 & 3.379 & 1.338\\
 \hline
\end{tabular}
\end{center}
\end{table*}

\subsection{Positional information}
\label{ssec:postion}

\begin{figure}[htb]
\includegraphics[width=8.5cm]{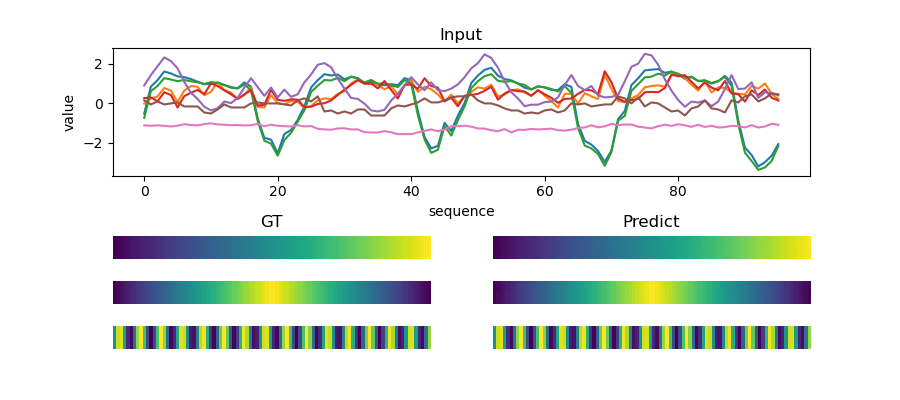}
\caption{Results of positional information prediction. The input data consists of 7-channel, 96-sequence data in test set of ETTh1. The tests include linear index, gradation, and sinusoidal signals. The GT is target data, and the predict is predicted data.}
\label{fig:position}
\end{figure}

\begin{table}[h!]
\begin{center}
\caption{Quantitative evaluation about position index}
\label{table:position}
\begin{tabular}{ c|c|c|c } 
 \hline
 Method & Linear index & Gradation & Sinusoid signal \\ 
 \hline
 MAE & 0.0017 & 0.0017 & 0.0019 \\
 \hline
\end{tabular}
\end{center}
\end{table}

In this sub-section, we show that the ARMA block contains positional information. First, we use only CNN part of the frozen model that is trained in the ETTh1 dataset. Then add a single linear layer as Position Encoding Module(PEM). We used ETTh1 data that sequence is 96 and channel is 7 to train PEM. We train only PEM using ETTh1 training parts for each indexing data(linear index, gradation, sinusoid signal) and predict indexing using ETTh1 test data. The result is shown in Fig.\ref{fig:position}. and Table \ref{table:position}. This shows that each filtered result by CNN contains positional information.

\subsection{Quantitative Results}
\label{ssec:result}

In Table \ref{table:result}, we extensively evaluate each method on nine benchmarks. The results show that the best performance was achieved in the four cases of Exchange, National Visits Influenza-like Illnesses(ILI), ETTh2, and ETTm2, whilst the remaining parts demonstrated performance similar to that of existing transformer models. This indicates that the ARMA block is sufficiently capable of time series forecasting.

\begin{figure}[htb]
\includegraphics[width=8.5cm]{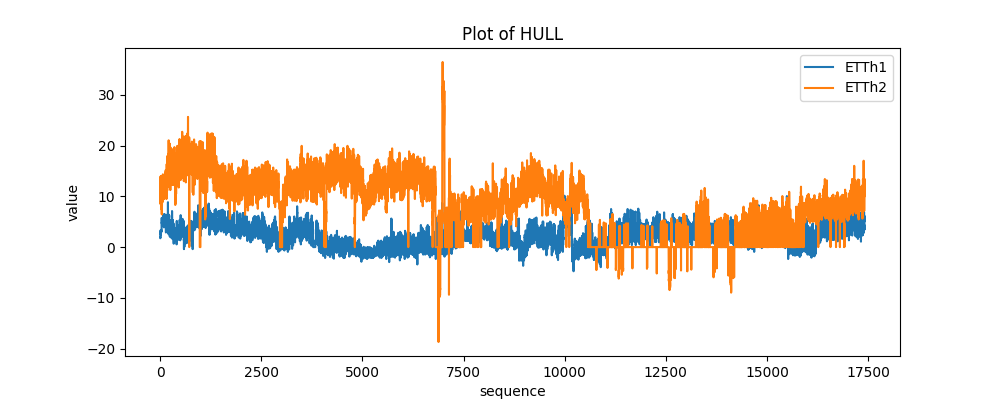}
\caption{Output of values for the HULL column in ETTh1 and ETTh2}
\label{fig:hull_figure}
\end{figure}

In contrast, on datasets characterized by trend shifts and distributional non-stationarity—notably Exchange, ILI, ETTh2, and ETTm2 \cite{liu2023adaptive}—the ARMA block achieves the best or highly competitive performance, surpassing both transformer and linear baselines in multiple forecasting horizons. This aligns with the intuition that the AR component captures evolving long-term trends, whereas the MA component corrects high-frequency residuals, yielding robustness under shifting data regimes. Fig. \ref{fig:hull_figure} further illustrates the presence of trend changes (e.g., in ETTh2) that favor this decomposition.

Overall, the experiments confirm that the ARMA block combines simplicity, efficiency, and robustness, making it a practical module for real-world long-term time series forecasting where both adaptability and low complexity are crucial. These results highlight that the ARMA block provides a rare combination of robustness to trend shifts and computational simplicity, making it well-suited for practical deployment in large-scale forecasting pipelines.

\subsection{Ablation study}
\label{ssec:ablation}

To investigate the effectiveness of the ARMA block, comparative experiments were conducted using a block employing a simple CNN. As the ARMA block consists solely of two CNNs—one for the AR component and one for the MA component—it effectively becomes a block composed of a simple CNN when one of these is excluded. This experiment was performed only for the case where the MA component was excluded. The MA component of the ARMA block cannot operate independently; consequently, it was excluded from this ablation study.

\begin{table}[h!]
\begin{center}
\caption{Quantitative evaluation for ablation studies}
\label{table:ablation}
\begin{tabular}{ c|c|c } 
 \hline
 Method & ARMA block & CNN block \\ 
 \hline
 MSE & 1.353 & 1.699 \\
 MAE & 0.776 & 0.914 \\
 \hline
\end{tabular}
\end{center}
\end{table}

\begin{figure}[htb]
\includegraphics[width=8.5cm]{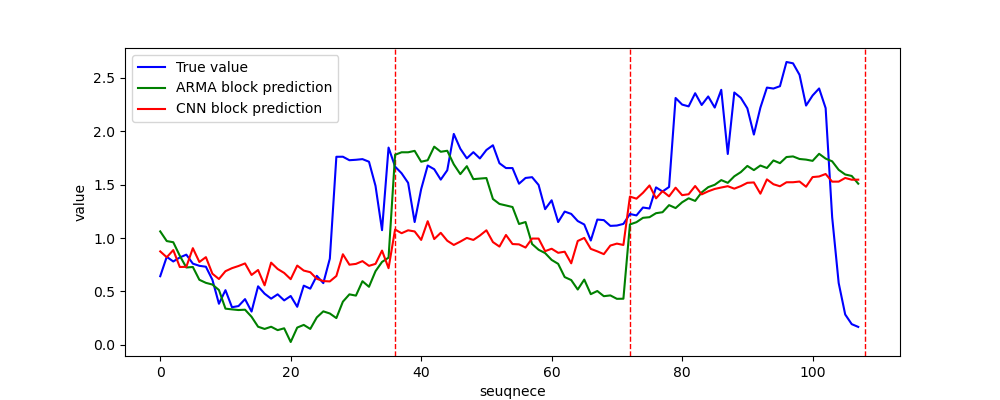}
\caption{The result of prediction from CNN and ARMA block in ILI dataset. The blue line is true value, the green line is prediction value of ARMA block and the red line is prediction value of simple CNN. The red vertical line is forecasting sequence step point.}
\label{fig:ablation_figure}
\end{figure}

When training on the ILI dataset under identical conditions, the quantitative results in Table \ref{table:ablation} reveal that the case employing the ARMA block exhibited lower error. This difference in error can be observed in Fig. \ref{fig:ablation_figure}. For the CNN, while it follows the main trend, it lacks a component to assist with the details, meaning it cannot predict finer trends. This point can be seen as evidence that the learning proceeds in the direction intended by the ARMA block from the outset, with the MA correcting information regarding the finer details.

\section{Conclusion}
\label{sec:conclusion}

We presented the ARMA block, a simple forecasting module that combines efficiency, robustness, and adaptability. By capturing both long-term trends and high-frequency residuals, the block is particularly effective on datasets with trend shifts, achieving state-of-the-art results in several cases. While its accuracy does not always surpass that of the most recent linear baselines, it achieves competitive results against both linear and transformer-based models across diverse datasets, demonstrating that the design remains robust despite its simplicity. Importantly, it encodes positional information without explicit embeddings, reducing model complexity. These findings suggest its utility as a lightweight component in real-world forecasting pipelines and as a potential alternative to structured state-space modules like Mamba.

\bibliographystyle{IEEEbib}
\bibliography{strings,refs}

\begin{thebibliography}{10}

\bibitem{vaswani2017attention}
Ashish Vaswani, Noam Shazeer, Niki Parmar, Jakob Uszkoreit, Llion Jones, Aidan~N Gomez, {\L}ukasz Kaiser, and Illia Polosukhin,
\newblock ``Attention is all you need,''
\newblock {\em Advances in neural information processing systems}, vol. 30, 2017.

\bibitem{dosovitskiy2020image}
Alexey Dosovitskiy, Lucas Beyer, Alexander Kolesnikov, Dirk Weissenborn, Xiaohua Zhai, Thomas Unterthiner, Mostafa Dehghani, Matthias Minderer, Georg Heigold, Sylvain Gelly, Jakob Uszkoreit, and Neil Houlsby,
\newblock ``An image is worth 16x16 words: Transformers for image recognition at scale,''
\newblock in {\em International Conference on Learning Representations}, 2021.

\bibitem{gu2023mamba}
Albert Gu and Tri Dao,
\newblock ``Mamba: Linear-time sequence modeling with selective state spaces,''
\newblock {\em arXiv preprint arXiv:2312.00752}, 2023.

\bibitem{gu2024mamba}
Albert Gu and Tri Dao,
\newblock ``Mamba: Linear-time sequence modeling with selective state spaces,''
\newblock in {\em First Conference on Language Modeling}, 2024.

\bibitem{zhou2021informer}
Haoyi Zhou, Shanghang Zhang, Jieqi Peng, Shuai Zhang, Jianxin Li, Hui Xiong, and Wancai Zhang,
\newblock ``Informer: Beyond efficient transformer for long sequence time-series forecasting,''
\newblock in {\em Proceedings of the AAAI conference on artificial intelligence}, 2021, vol.~35, pp. 11106--11115.

\bibitem{wu2021autoformer}
Haixu Wu, Jiehui Xu, Jianmin Wang, and Mingsheng Long,
\newblock ``Autoformer: Decomposition transformers with auto-correlation for long-term series forecasting,''
\newblock {\em Advances in neural information processing systems}, vol. 34, pp. 22419--22430, 2021.

\bibitem{zhou2022fedformer}
Tian Zhou, Ziqing Ma, Qingsong Wen, Xue Wang, Liang Sun, and Rong Jin,
\newblock ``Fedformer: Frequency enhanced decomposed transformer for long-term series forecasting,''
\newblock in {\em International conference on machine learning}. PMLR, 2022, pp. 27268--27286.

\bibitem{shi2024time}
Xiaoming Shi, Shiyu Wang, Yuqi Nie, Dianqi Li, Zhou Ye, Qingsong Wen, and Ming Jin,
\newblock ``Time-moe: Billion-scale time series foundation models with mixture of experts,''
\newblock in {\em The Thirteenth International Conference on Learning Representations}, 2025.

\bibitem{liu2024mixture}
Xu~Liu, Juncheng Liu, Gerald Woo, Taha Aksu, Chenghao Liu, Silvio Savarese, Caiming Xiong, and Doyen Sahoo,
\newblock ``Mixture of experts for time series foundation models,''
\newblock in {\em NeurIPS Workshop on Time Series in the Age of Large Models}, 2024.

\bibitem{zeng2023transformers}
Ailing Zeng, Muxi Chen, Lei Zhang, and Qiang Xu,
\newblock ``Are transformers effective for time series forecasting?,''
\newblock in {\em Proceedings of the AAAI conference on artificial intelligence}, 2023, vol.~37, pp. 11121--11128.

\bibitem{islam2020much}
Md~Amirul Islam, Sen Jia, and Neil~DB Bruce,
\newblock ``How much position information do convolutional neural networks encode?,''
\newblock {\em arXiv preprint arXiv:2001.08248}, 2020.

\bibitem{box2015time}
George~EP Box, Gwilym~M Jenkins, Gregory~C Reinsel, and Greta~M Ljung,
\newblock {\em Time series analysis: forecasting and control},
\newblock John Wiley \& Sons, 2015.

\bibitem{lai2018modeling}
Guokun Lai, Wei-Cheng Chang, Yiming Yang, and Hanxiao Liu,
\newblock ``Modeling long-and short-term temporal patterns with deep neural networks,''
\newblock in {\em The 41st international ACM SIGIR conference on research \& development in information retrieval}, 2018, pp. 95--104.

\bibitem{kim2021reversible}
Taesung Kim, Jinhee Kim, Yunwon Tae, Cheonbok Park, Jang-Ho Choi, and Jaegul Choo,
\newblock ``Reversible instance normalization for accurate time-series forecasting against distribution shift,''
\newblock in {\em International conference on learning representations}, 2021.

\bibitem{liu2023adaptive}
Zhiding Liu, Mingyue Cheng, Zhi Li, Zhenya Huang, Qi~Liu, Yanhu Xie, and Enhong Chen,
\newblock ``Adaptive normalization for non-stationary time series forecasting: A temporal slice perspective,''
\newblock {\em Advances in Neural Information Processing Systems}, vol. 36, pp. 14273--14292, 2023.

\end{thebibliography}

\end{document}